%% file: main.tex
\pdfoutput=1
\documentclass[11pt]{article}

\usepackage[final]{mtsummit25}
\usepackage{copyright}

\usepackage{times}
\usepackage{latexsym}

\usepackage[T1]{fontenc}
\usepackage[utf8]{inputenc}

\usepackage{microtype}

\usepackage{inconsolata}

\usepackage{graphicx}
\usepackage[mathscr]{euscript}
\usepackage[ruled,linesnumbered]{algorithm2e}
\usepackage{booktabs}

\title{Optimizing the Training Schedule of Multilingual NMT\\using Reinforcement Learning}

\author{
 \textbf{Alexis Allemann\textsuperscript{1}},
 \textbf{\`{A}lex R. Atrio\textsuperscript{2,*}}, 
 \textbf{Andrei Popescu-Belis\textsuperscript{1,3}}
\\
\\
 \textsuperscript{1}HEIG-VD~/~HES-SO, 1401 Yverdon-les-Bains, Switzerland\\
 \textsuperscript{2}Pi School, Rome, Italy\\
 \textsuperscript{3}EPFL, 1015 Lausanne, Switzerland
\\
 \small{
   \textbf{Correspondence:} \href{mailto:andrei.popescu-belis@heig-vd.ch}{andrei.popescu-belis@heig-vd.ch}
 }
}

\begin{document}
\maketitle
\begin{abstract}
Multilingual NMT is a viable solution for translating low-resource languages (LRLs) when data from high-resource languages (HRLs) from the same language family is available.  However, the training schedule, i.e.\ the order of presentation of languages, has an impact on the quality of such systems.  Here, in a many-to-one translation setting, we propose to apply two algorithms that use reinforcement learning to optimize the training schedule of NMT: (1)~Teacher-Student Curriculum Learning and (2)~Deep Q~Network.  The former uses an exponentially smoothed estimate of the returns of each action based on the loss on monolingual or multilingual development subsets, while the latter estimates rewards using an additional neural network trained from the history of actions selected in different states of the system, together with the rewards received.  On a 8-to-1 translation dataset with LRLs and HRLs, our second method improves BLEU and COMET scores with respect to both random selection of monolingual batches and shuffled multilingual batches, by adjusting the number of presentations of LRL vs.\ HRL batches.
\end{abstract}

\section{Introduction}
\label{sec:introduction}

Multilingual neural machine translation (NMT)is particularly effective to enable the translation of low-resource languages (LRLs) when they are accompanied, in the training data, by related high-resource languages (HRLs) \citep{gu-etal-2018-universal,neubig-hu-2018-rapid}.  Including HRLs in the training data reduces the chance of overfitting to the LRLs and improves translation quality.

Many-to-one NMT systems can be trained either with monolingual or with multilingual batches. Monolingual batches include a single language on the source side, while multilingual batches have their source side sampled from several source languages.  Using multilingual batches helps avoiding catastrophic forgetting \citep{jean2019adaptive}, but the mixture of languages in each batch may be ineffective at early stages of training.  Here, we focus on monolingual batches, as they enable us to define the training schedule of a NMT system as the order of presentation of languages, but we also compare our results to those obtained with multilingual batches.

We propose to use reinforcement learning (RL) to optimize the training schedule of many-to-one NMT systems, i.e.\ to improve the training process and the resulting system compared to a fixed sampling strategy.  We enable our systems to select the source language of the batch at each training step, based on a learned estimate of the model's competence on each language in terms of loss on a development set.  Unlike fixed strategies, such as training on the hardest language, we leverage RL to let the model find better strategies.

We make the following contributions:\footnote{Source code is made available at \href{https://github.com/alexis-allemann/OpenNMT-py/tree/curriculum_learning}{https://github.com/alexis-allemann/OpenNMT-py/tree/curriculum\_learning}.}

\begin{itemize} \setlength{\itemsep}{0pt}
    \item We apply the Teacher-Student Curriculum Learning algorithm \citep{Matiisen-2017} to NMT by modeling the expected return as the smoothed loss of the NMT system over a development set.
    \item Based on the Deep Q Network algorithm (DQN) \citep{Mnih-2013}, we design a RL-based model in which the expected rewards are generated by an auxiliary neural network trained in parallel with the NMT system.  
    \item We perform experiments on a dataset with four language families on the source side, with one HRL and one LRL for each family \citep{neubig-hu-2018-rapid}; the target language is English.
    \item DQN outperforms in terms of BLEU and COMET scores previous training schedules used for multilingual NMT: monolingual minibatches sampled equally, or in proportion of each language, or multilingual batches.
    \item Algorithms are robust to the setting of hyper-parameters, and increase the proportions of LRLs in the training schedule from less than 1\% to at least 4\% while decreasing those of HRLs.
\end{itemize}


\section{Related Work}
\label{sec:related-work}

\paragraph{Curriculum learning.}
In many applications of machine learning, the order of presentation of items from the training set may influence the outcome of the training, i.e.\ the quality of the final model, or the training speed.
For instance, presenting items by increasing levels of difficulty is often beneficial, an approach known as \emph{curriculum learning} \citep{wang_survey_2021}.
The difficulty can be measured directly on the data, or it can be inferred from the observed competence of the model during training, an approach known as \emph{self-paced learning} \citep{kumar2010self,jiang2015self}. 
The competence of a model can be estimated intrinsically, e.g.\ from its loss values on a subset of the data, or extrinsically, by using a teacher model that observes the behavior of the target model, called `student' \citep{Matiisen-2017}.  Competence can be used by the teacher model to adjust the training schedule of the student model.
In the case of systems that can perform several tasks, the training schedule consists of the selection of tasks and related data.

When the teacher model is in charge of the training schedule of the student, it may use reinforcement learning (RL), with the student model playing the role of the environment \citep{shen-zhao-rl4nlp-2024}.
RL has proved particularly useful at training large language models to follow instructions \citep{ouyang2022traininglanguagemodelsfollow}, initially using PPO \citep[Proximal Policy Optimization,][]{DBLP:journals/corr/SchulmanWDRK17} and then other algorithms \citep{rafailov2023directpreferenceoptimizationlanguage,ethayarajh2024ktomodelalignmentprospect}, 
but these methods are not designed to optimize training schedules.
While it is possible to use curriculum learning to train RL-based models \citep{NarvekarCLforRL2020}, e.g.\ by presenting them with increasingly difficult problems, we focus here on the use of RL to train a teacher model, in the field of multilingual NMT. 

\paragraph{Training schedules for NMT.}
Optimizing the training schedule of an NMT system depends in particular on its architecture.  For a system with a single input language and domain, the training sentences can be presented by order of estimated difficulty, or by order of translation reliability or noisiness.  When multiple domains must be considered, additional decisions must be made about which domain to use first, or how to mix them based on sizes of available data.  Similar decisions must be made if there are multiple input languages, as in our case, or if one must train a multi-task system including NMT along with other tasks such as language modeling.  We briefly review here previous work along these lines. 

\paragraph{Static scheduling in multilingual NMT.}
\citet{neubig-hu-2018-rapid} study the upsampling of the LRL data when building minibatches, and observe that keeping the original proportions of HRL and LRL performs marginally better.  However, \citet{johnson-etal-2017-googles} and \citet{aharoni-etal-2019-massively} sample each batch uniformly from a concatenation of all language pairs.  \citet{arivazhagan2019massively} compare simple concatenation with uniform balancing, and observe better results for LRLs when using temperature-based upsampling, which was favored afterwards \citep{conneau-etal-2020-unsupervised,tang-etal-2021-multilingual}.  

The translation capabilities of large language models (LLMs) have also been explored: \citet{zhu-etal-2024-multilingual} compares several recent LLMs and shows that they can achieve state-of-the-art results when translating HRLs, but highlights their limitations in translating LRLs compared to NMT models.  One of the leading open-weights LLMs for MT, Tower Instruct \citep{alves2024toweropenmultilinguallarge}, is fine-tuned on a large set of translation-related tasks in 10 HRLs, with no particular scheduling of the fine-tuning data, and no reinforcement learning.

\paragraph{Curriculum learning in monolingual NMT.} 
Self-pacing has been used in NMT at the sample level, for instance by estimating learning confidence as the variance across dropout runs, with better performance and faster convergence compared to human-designed schedules \citep{wan-etal-2020-self}.  Similarly, \citet{liu-etal-2020-norm} design a self-paced curriculum based on the norm of a token's embedding.  \citet{zhang2018empirical} adopt a probabilistic view of curriculum learning and improve the convergence time of a DE-EN NMT system at no loss in translation quality, but no gain either; moreover, they note a high sensitivity to hyperparameter settings.  
\citet{platanios-etal-2019-competence} propose a scheduling criterion combining the difficulty of samples and the competence of the NMT model, the latter estimated as a linear or square root function of the number of steps. This reduces training time by up to 70\% and improves BLEU scores by 1--2 points on three different language pairs. 
\citet{wang_denoising_2018} extend domain-specific data selection methods to denoise NMT training, which significantly improves NMT performance on noisy data.
\citet{wang_learning_2020} introduce a method for multi-domain data selection in NMT, using instance-level domain-relevance features and an automated training curriculum to enhance performance across multiple domains. 

\paragraph{Curriculum learning in multilingual NMT.}
\citet{jean2019adaptive} compare adaptively upsampling a language depending on various criteria, observing best results on LRLs when dynamically changing the norm of the gradient. 
\citet{wang-etal-2020-balancing} adaptively balance the languages by learning their weights from the model's competence on a development set.
\citet{zhang-etal-2021-competence-based} design a dynamic sampling strategy which measures per-language competence but also evaluates LRL competence through a related HRL's competence.
\citet{wu-etal-2021-uncertainty} also balance the data dynamically, but measure a model's uncertainty as the variance over several runs of Monte Carlo dropout. 
Estimates of competence using the evolution of the loss of the NMT system have been proposed by \citet{zaremoodi-haffari-2019-adaptively}, who use its absolute value, by \citet{xu-etal-2020-dynamic}, who use its relative decrease, and by \citet{atrio-etal-2024-selfpaced}, who use Kullback-Leibler divergence between consecutive states of the weights of an entire Transformer network.

\paragraph{RL-based curriculum learning in NMT.}
In the field of machine translation, \citet{kumar-etal-2019-reinforcement} propose a RL framework utilizing Q-Learning to automatically learn an optimal curriculum for heterogeneous data, matching state-of-the-art hand-designed curricula.  \citet{zhao2020reinforced} introduce a RL-based data selection framework using Deterministic Actor-Critic to improve pre-trained NMT models by re-selecting influential samples from the original training set. \citet{kreutzer2021banditsdontfollowrules} use a multi-armed bandit to dynamically select training data, thus optimizing NMT model performance across different domains, data qualities, and language pairs without manual schedule design.

\paragraph{Other applications of RL to NMT.}
In machine translation, RL methods were employed by \citet{edunov-etal-2018-classical} to tackle the discrepancy between token-level likelihood optimization during training and corpus-level evaluations using metrics like BLEU, and to reduce exposure bias in autoregressive sequence generators \citep{ranzato2016sequenceleveltrainingrecurrent,wang-sennrich-2020-exposure,wu-etal-2018-study}.  \citet{Kiegeland_2021} emphasize the importance of exploration strategies, reward scaling, and reward function design for improving translation quality, particularly with respect to domain adaptation.  To enhance the effectiveness of RL in NMT, \citet{yehudai_reinforcement_2022} show the importance of reducing the size and dimensionality of the action space.  \citet{wang2024esrl} introduce efficient sampling-based RL techniques for sequence generation models, with a strong focus, however, on instruction tuning of LLMs.

\section{Two RL Algorithms for Optimizing Training Schedules}
\label{sec:rl4tso}
In the RL framework, an agent observes the state $S_t$ of the environment at each time step $t$, selects an action $A_t$ based on its policy $\pi$, executes the action, and receives a reward $R_t$ from the environment.  Using the observed states, actions, and rewards, the goal is to learn an optimal policy, i.e.\ one that maximizes the cumulative reward over time.  Bandit problems are those where the agent selects actions without considering state transitions.

In this study, as we use for training only monolingual batches, the set of possible actions $\mathscr{A}$ is simply the set of source languages.  The states $\mathscr{S}$ of the system are the values of the parameters of the neural network and of the optimizer. However, these are too numerous to be sensibly observed at each step. Drawing inspiration from \citet{wu-etal-2018-reinforced}, we compute the current state of the model as the vector of cross-entropy loss values obtained from the NMT system over a development batch of sentences.\footnote{Alternatively, MT-specific metrics such as BLEU or COMET could be used instead of the cross-entropy loss, but computing them is more costly, therefore we do not use them here.}  
We use the score $X_t$ of the model at time step $t$ to compute the reward $R_t$ as the decrease of the loss of the NMT system between the last two time steps: $R_t = X_t - X_{t-1}$.  The loss values are computed on a development minibatch of data selected from the current language, or, in some experiments, on a multilingual minibatch. 


\subsection{TSCL Algorithm for NMT}

Our first proposal is an adaptation to NMT of the Teacher-Student Curriculum Learning (TSCL) algorithm, a bandit method introduced by \citet{Matiisen-2017}, who use it to add decimal numbers or to navigate Minecraft mazes.  The gist of our adaptation of TSCL for multilingual NMT is represented in Figure~\ref{fig:rl-tscl-mnmt}, and the full algorithm is given in Appendix~\ref{sec:appendix-tscl}.

\begin{figure}[ht]
    \centering
    \includegraphics[width=1.05\linewidth]{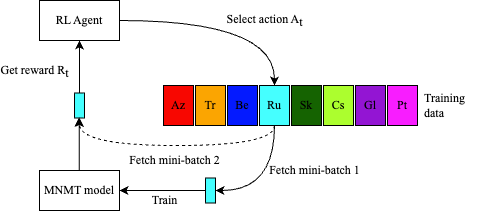}
    \caption{TSCL algorithm for NMT: relationship between the RL agent and the NMT system.}
    \label{fig:rl-tscl-mnmt}
\end{figure}

The action $A_t$ of the system at time step $t$ is the selection of a batch from a specific source language for training in the next step.  The reward $R_t$ of the action is the decrease of the negative cross-entropy loss of the model, $X_t$, computed on a batch of 8k tokens from the current source language, with respect to the value $X_{t'}$ computed at the latest previous time step with the same language.  Formally, $R_t = X_{t} - X_{t'}$ where $t' = \mathrm{max}\{s : s < t \mathrm{\ and\ } A_s = A_t\}$.  The expected return $Q_t$ of the action is the exponentially weighted moving average of the rewards for the respective source language.

In some experiments, we start with a warm-up phase, a period during which all HRLs are randomly explored while the learning rate of the NMT model increases.  Rewards of the RL agent only start to play a role after the warm-up phase, when the learning rate starts decreasing.
In experiments without warm-up, each action is executed once at the beginning of the training, so that the model initiates training on the language that provides the highest reward from the start.  

Additionally, to strike a balance between exploration and exploitation, we use an $\epsilon$-greedy policy with a fixed value of $\epsilon$.  The action with the highest expected return is selected with probability $1 - \epsilon$, but with a small probability $\epsilon$ a random action is selected.  In experiments with a warm-up period, this policy only starts after this period.


\begin{figure*}[ht]
    \centering
    \includegraphics[width=0.9\linewidth]{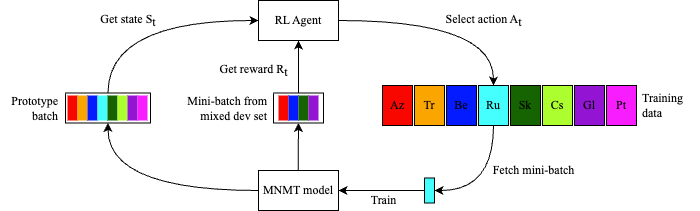}
    \caption{DQN algorithm for NMT: relationship between the RL agent (Q network) and the NMT model.}
    \label{fig:rl-dqn-mnmt}
\end{figure*}

\subsection{DQN Algorithm for NMT}
\label{sec:dqn-algorithm}

The Deep Q Network (DQN) algorithm \citep{Mnih-2013} uses a neural network to approximate the Q-function that represents the expected reward of an action in a given state. The algorithm iteratively updates the parameters of this network to minimize the difference between the predicted Q-values and the desired Q-values obtained from the target system.  Moreover, DQN enables experience replay by storing past experiences in a replay buffer and sampling them randomly during the training of the Q network, a feature that was shown to improve training.

Our application of DQN to multilingual NMT is illustrated in Figure~\ref{fig:rl-dqn-mnmt}, and the full algorithm is given in Appendix~\ref{sec:appendix-dqn}.  The RL agent is the Q network, a feed-forward neural network with \emph{tanh} activation functions.  Its input is the state of the NMT model: specifically, each value in the input layer represents the cross-entropy loss of the NMT model over a batch of 10 sentences from a specific language. Thus, an input vector of size 200 corresponds to a prototype batch of 2,000 sentences, with 250 sentences from each of the 8 source languages. 
The input layer is followed by two hidden layers of size 512 and by an output layer with 8 units, corresponding to the possible actions (selection of a source language for the next training step).  The Q network is trained with the RMSProp 
optimizer\footnote{A variant of stochastic gradient descent, proposed by G.~Hinton, which adapts the learning rate for each parameter based on recent gradient averages \citep{Ruder-2016}.} 
and the Huber Loss \citep{10.1214/aoms/1177703732}, a loss function that reduces the influence of extreme values, to mitigate the issue of outliers during training.

At each timestep $t$, the RL agent retains a new transition in its experience replay buffer. A transition consists of the previous state of the system $S_{t-1}$, the selected action $A_{t-1}$, the obtained reward $R_t$, and the current state of the system $S_t$. 
These transitions are used to train the Q network so that it predicts the action with the best estimated reward given the state of the NMT model.

We use an $\epsilon$-greedy policy to balance between exploring actions and exploiting the Q network, like for TSCL.  During the warm-up period, which is always applied to DQN, actions are randomly selected, but after it, actions are selected by the Q network with a probability of $1 - \epsilon$ or thy are randomly selected with a probability of $\epsilon$.  However, unlike TSCL, we follow \citet{kumar-etal-2019-reinforcement} and start with $\epsilon = 1$ during warm-up, then gradually decrease this value at the end of warm-up to a minimum of 0.01 after 50k steps.  This allows the network to randomly explore actions during the warm-up period before exploiting the learned knowledge more and more.  The schedule of $\epsilon$ is represented in Figure~\ref{fig:epsilon-decay} of Appendix~\ref{app:epsilon-scheduling}.


\section{Experimental Settings}
\label{sec:settings}

\paragraph{Data.}
Experiments were conducted using a subset of the multilingual TED corpus collected by \citet{qi-etal-2018-pre}, with four HRLs and four LRLs.\footnote{\href{https://github.com/neulab/word-embeddings-for-nmt}{github.com/neulab/word-embeddings-for-nmt}}  For comparability with prior  research on multilingual NMT \citep{neubig-hu-2018-rapid,wang2019multilingual,zhang-etal-2021-competence-based}, we consider a 8-to-1 translation task with English as the target language.  We are especially interested in the translation quality of the four LRLs of the dataset: Belarusian (\textsc{Be}), Azerbaijani (\textsc{Az}), Galician (\textsc{Gl}) and Slovak (\textsc{Sk}), which are respectively paired with a HRL from the same family: Russian (\textsc{Ru}), Turkish (\textsc{Tr}), Portuguese (\textsc{Pt}) and Czech (\textsc{Cs}).  Three language families are thus represented (Romance, Slavic and Turkic) but all scripts are Latin-based.

The numbers of sentences of the training and testing sets for each of the 8 languages are shown in Table~\ref{tab:data}.  These numbers show that the distinction of LRLs vs.\ HRLs made in previous studies is to some extent arbitrary.  Indeed, there are fewer \textsc{Pt} sentences (considered nevertheless as a HRL with respect to \textsc{Gl}) than \textsc{Sk} sentences (considered as a LRL with respect to \textsc{Cs}).  

\begin{table}[ht]
    \centering
    \begin{tabular}{c|ccc}
        \toprule
            \textbf{Language} & train & dev & test \\
        \midrule
            \textsc{Az} & 5.9k (0.95\%) & 671 & 903 \\
            \textsc{Be} & 4.5k (0.72\%) & 248 & 664 \\ 
            \textsc{Gl} & 10.0k (1.60\%) & 682 & 1.0k \\  
            \textsc{Sk} & 61.5k (9.79\%) & 2.2k & 2.4k \\ 
        \midrule 
            \textsc{Tr} & 182k (29.07\%) & 4.0k & 5.0k \\
            \textsc{Ru} & 208k (33.21\%) & 4.8k & 5.5k \\
            \textsc{Pt} & 51.8k (8.25\%) & 1.2k & 1.8k \\
            \textsc{Cs} & 103k (16.42\%) & 3.5k & 3.8k \\
        \bottomrule
    \end{tabular}
    \caption{Numbers of sentences for LRLs and HRLs.}
    \label{tab:data}
\end{table}

\paragraph{Preprocessing.}
The original data is already tokenized into words.  We use Byte Pair Encoding (BPE) for subword extraction and vocabulary construction \citep{sennrich-etal-2016-neural}.\footnote{\href{https://github.com/rsennrich/subword-nmt}{github.com/rsennrich/subword-nmt}}  A vocabulary of 32k subwords is generated over a multilingual corpus obtained by combining 10k random lines from the training data of each language, with upsampling for \textsc{Az} and \textsc{Be} which have fewer than 10k lines. For source language identification by the NMT model, each sentence is prefixed with a language tag.

\paragraph{NMT Models.}
We experiment with Transformer models from the OpenNMT-py library version 3.4.3 \citep{klein-etal-2017-opennmt}.\footnote{\href{https://github.com/OpenNMT/OpenNMT-py/releases/tag/v3.4.3}{github.com/OpenNMT/OpenNMT-py}}  All models are trained for 150k steps. The hyperparameter values are the default ones from the Transformer-Base model \citep{NIPS2017_3f5ee243}: 6 layers for the encoder and 6 for the decoder, 8 attention heads, label smoothing of 0.1, hidden layers with 512 units, and feedforward networks with 2,048 units. The Adam optimizer \citep{kingma-adam2014} is used.  Following \citet{atrio-popescu-belis-2022-interaction}, we use a batch size of 8k tokens and the regularization parameters are: dropout rate of 0.3, scaling factor of 10, and gradients are re-normalized if their norm exceeds~5.  In experiments with warm-up, there are 16k steps during which the learning rate increases from 0 to its maximum.

\paragraph{RL Agents.}
Several hyperparameters must be set for RL Agents.  Their default values are given here, while the behavior of the systems when these values are modified are studied in Section~\ref{sec:parameter-setting}. 

The TSCL algorithm is run with a smoothing coefficient $\alpha=0.1$.
The warm-up period is 16k steps, during which batches from HRLs are presented in a random order.  For the $\epsilon$-greedy policy, $\epsilon=0.1$.  These values correspond to those used by \citet{Matiisen-2017}.

The DQN algorithm is also run with a warm-up period of 16k steps on HRLs only.  Unlike TSCL, a new action is selected every 10 steps, and not at every step, to reduce computing time, with no significant differences in observed results.  The Q network underlying the RL agent has an input layer with 200 units, two fully connected subsequent layers with 512 units each, and an output layer with 8 units.  As explained in Section~\ref{sec:dqn-algorithm}, each value in the input layer corresponds to the cross-entropy loss of the NMT model over a batch of 10 sentences from a specific language. 

The training of the Q network has a learning rate $lr = 2.5\mathrm{e}-4$ and a soft update smoothing coefficient $\tau = 0.005$.  The discount factor, which influences the importance of future rewards in the agent's decision-making process, is $\gamma = 0.99$.\footnote{This parameter is defined in Section~2 of \citet{Mnih-2013} and is implicit in line 19 of our Algorithm~\ref{alg:DQN}.}
The experience replay buffer has minimal/maximal sizes of 1k/10k.  These values are those used by \citet{kumar-etal-2019-reinforcement}.  

\paragraph{Evaluation Metrics.}
Translation quality is measured using the BLEU and COMET metrics.  BLEU scores are computed with the SacreBLEU library \citep{post-2018-call}.\footnote{\href{https://github.com/mjpost/sacrebleu}{github.com/mjpost/sacrebleu}, signature: \texttt{nrefs:1|case: mixed|eff:no|tok:13a|smooth:exp|version:2.3.1.}} 
COMET scores are computed using the \texttt{wmt22-comet-da} model \citep{rei-etal-2022-comet}.\footnote{\href{https://huggingface.co/Unbabel/wmt22-comet-da}{huggingface.co/Unbabel/wmt22-comet-da}}  Scores are computed using a rolling ensemble of four checkpoints. The best ensemble in terms of average BLEU score on the LRLs development sets is used to translate the test set.


\section{Results and Analysis}
\label{sec:results}

\subsection{Baselines} 

We compare TSCL and DQN to baseline training schedules in which source languages are selected randomly at each step, either with a uniform distribution ($P=1/8$ for each language) or with a distribution that is proportional to the number of sentences of the respective language in the training data -- hence between 0.95\% for \textsc{Az} and 33.21\% for \textsc{Ru}, as shown in Table~\ref{tab:data}.  Moreover, a warm-up period of 16k on HRLs can be used or not.  This results in four baseline schedules, shown in the first four lines of Tables~\ref{tab:results-baselines-rl-lrl}, \ref{tab:results-baselines-rl-hrl} and~\ref{tab:results-baselines-rl-avg}.  While these baselines and the TSCL and DQN algorithms use monolingual batches, a fifth baseline uses multilingual shuffled ones, with sentences drawn randomly from the source languages in proportion to their frequency, and a warm-up period of 16k on HRLs. Shuffled batches were found to perform particularly well on this dataset \citep{neubig-hu-2018-rapid,atrio-etal-2024-selfpaced}.

\begin{table*}[ht!]
    \centering
    \begin{tabular}{ll|cc|cc|cc|cc}
        \toprule
            \textbf{Training} & \textbf{Warm} & \multicolumn{2}{c|}{\textsc{Az} $\rightarrow$ \textsc{En}} & \multicolumn{2}{c|}{\textsc{Be} $\rightarrow$ \textsc{En}} & \multicolumn{2}{c|}{\textsc{Sk} $\rightarrow$ \textsc{En}} & \multicolumn{2}{c}{\textsc{Gl} $\rightarrow$ \textsc{En}} \\
            \textbf{schedule} & \textbf{up} & \textsc{Bleu} & \textsc{Comet} & \textsc{Bleu} & \textsc{Comet} &  \textsc{Bleu} & \textsc{Comet}  & \textsc{Bleu} & \textsc{Comet} \\    
        \midrule
            Uniform & no & 13.86 & 62.99 & 19.75 & 60.80 & 32.85 & 75.09 & 31.14 & 72.08 \\
            Proportional & no & \textbf{15.82} & 65.66 & 19.81 & 61.85 & \textbf{35.11} & \textbf{76.51} & 31.07 & 72.65 \\
            Uniform & 16k & 15.42 & 65.19 & 20.29 & 62.13 & 34.11 & 76.45 & 32.74 & \textbf{73.84} \\
            Proportional & 16k & 15.14 & 65.70 & 19.22 & 61.48 & 34.97 & 76.26 & 31.79 & 72.70 \\
            Shuffled batch & 16k & 14.37 & 64.37 & 20.08 & 62.15 & 33.92 & 76.28 & 32.15 & 72.99 \\
            \cmidrule(l){1-2} \cmidrule(r){3-10}
            TSCL & 16k & 14.89 & 65.04 & 20.10 & 61.96 & 34.35 & 76.23 & 32.64 & 73.59 \\
            DQN & 16k & 15.62 & \textbf{65.86} & \textbf{21.11} & \textbf{62.82} & 34.54 & 76.15 & \textbf{33.02} & 73.73 \\
        \bottomrule
    \end{tabular}
    \caption{Results of TSCL and DQN compared to baselines on LRLs.} 
    \label{tab:results-baselines-rl-lrl}
\end{table*}

\begin{table*}[ht!] 
    \centering
    \begin{tabular}{ll|cc|cc|cc|cc}
        \toprule
            \textbf{Training} & \textbf{Warm} & \multicolumn{2}{c|}{\textsc{Tr} $\rightarrow$ \textsc{En}} & \multicolumn{2}{c|}{\textsc{Ru} $\rightarrow$ \textsc{En}} & \multicolumn{2}{c|}{\textsc{Cs} $\rightarrow$ \textsc{En}} & \multicolumn{2}{c}{\textsc{Pt} $\rightarrow$ \textsc{En}} \\
            \textbf{schedule} & \textbf{up} & \textsc{Bleu} & \textsc{Comet} & \textsc{Bleu} & \textsc{Comet} &  \textsc{Bleu} & \textsc{Comet}  & \textsc{Bleu} & \textsc{Comet} \\    
        \midrule
            Uniform & no & 27.26 & 75.41 & 26.95 & 72.18 & 30.87 & 74.30 & 39.76 & 78.41 \\
            Proportional & no & \textbf{29.40} & \textbf{77.48} & \textbf{28.14} & \textbf{74.04} & \textbf{32.47} & \textbf{75.79} & 37.98 & 76.56 \\
            Uniform & 16k & 28.29 & 76.72 & 27.53 & 73.32 & 31.76 & 75.60 & 41.32 & 79.57 \\
            Proportional & 16k & 28.97 & 77.25 & 27.96 & 73.53 & 32.16 & 75.38 & 38.64 & 76.47 \\
            Shuffled batch & 16k & 28.25 & 76.76 & 27.31 & 73.20 & 31.30 & 75.37 & 40.58 & 79.25 \\
            \cmidrule(l){1-2} \cmidrule(r){3-10}
            TSCL & 16k &  28.50 & 76.64 & 27.56 & 72.99 & 31.76 & 75.15 & \textbf{42.38}  & 79.52 \\
            DQN & 16k & 28.11 & 76.45 & 27.66 & 73.28 & 31.89 & 75.31 & 42.09 & \textbf{79.73} \\
        \bottomrule
    \end{tabular}
    \caption{Results of TSCL and DQN compared to baselines on HRLs.} 
    \label{tab:results-baselines-rl-hrl}
\end{table*}

\begin{table}[ht!] 
\centering
\begin{tabular}{ll|cc}
\toprule
\textbf{Training} & \textbf{Warm} & \multicolumn{2}{c}{\bf Average} \\
\textbf{schedule} & \textbf{up} & \textsc{Bleu} & \textsc{Comet} \\    
\midrule
Uniform        & no  & 27.81 & 71.40 \\
Proportional   & no  & 28.73 & 72.57 \\
Uniform        & 16k & 28.93 & 72.85 \\
Proportional   & 16k & 28.61 & 72.35 \\
Shuffled batch & 16k & 28.49 & 72.55 \\
\cmidrule(l){1-2} \cmidrule(r){3-4}
TSCL           & 16k & 29.02 & 72.64 \\
DQN            & 16k & \textbf{29.30} & \textbf{72.92} \\
\bottomrule
    \end{tabular}
    \caption{Macro-averages over all languages of the scores of TSCL and DQN compared to baselines.} 
    \label{tab:results-baselines-rl-avg}
\end{table}

\subsection{Translation Performance}
\label{sec:translation-performance}

The BLEU and COMET scores of the TSCL and DQN algorithms, in comparison to the baselines, are presented in Table~\ref{tab:results-baselines-rl-lrl} for the LRLs and in Table~\ref{tab:results-baselines-rl-hrl} for the HRLs.  The averages of BLEU and of COMET scores over the 8 languages are presented in Table~\ref{tab:results-baselines-rl-avg}, giving the same importance to each language, regardless of its frequency in the training data (macro-average).

The DQN algorithm outperforms on average all baselines, as well as the simpler TSCL algorithm, both in terms of BLEU and of COMET (Table~\ref{tab:results-baselines-rl-avg}).  The TSCL algorithm is second for BLEU, but third for COMET, slightly behind the uniform training schedule with warm-up.  Considering Table~\ref{tab:results-baselines-rl-lrl} with LRLs, we see that DQN often outperforms the other methods: it ranks first on COMET for \textsc{Az} and \textsc{Be} and second for \textsc{Gl} (but first on BLEU).  Moreover, DQN ranks first on COMET for \textsc{Pt}, as seen in Table~\ref{tab:results-baselines-rl-hrl}.  Therefore, DQN ranks first on three of the four least represented languages in the dataset.\footnote{As noted in Section~\ref{sec:settings}, the contrast between LRLs and HRLs made in previous work applies only within each pair of related languages.}
This shows that DQN improves learning of the LRLs at the price of a small degradation in HRLs, though still improving their macro-average.  
As for TSCL, although it is competitive on average with the baselines, it lags behind the best ones when it comes to individual languages.

The baseline that is most often ranked first is the one that selects batches in proportion to the frequency of the language in the training data, with no warm-up.  This has best BLEU and COMET scores on three HRLs (\textsc{Tr}, \textsc{Ru}, \textsc{Cs}) and one LRL$^\mathrm{10}$ (\textsc{Sk}), likely because each of these languages constitutes more than 10\% of the training data.
However, in this case, the NMT model struggles to learn LRLs because it does not see enough data from them.  As a result, when considering the macro-average, this baseline is slightly behind the one using warm-up on HRLs followed by selection of actions with uniform probability, which also has better COMET scores for \textsc{Be}, \textsc{Gl} and \textsc{Pt}.

Moreover, the DQN and TSCL algorithms are efficient in terms of convergence speed, defined as the number of steps needed to reach their best scores (the macro-average of BLEU on the LRLs of the development set).  As shown more fully in Table~\ref{tab:checkpoints} of Appendix~\ref{app:convergence-speed}, DQN reaches best performance after 52k steps, followed closely by TSCL and by the baseline with proportional batches (both at 60k).  The baseline with uniformly-drawn batches needs twice more steps to converge.


\begin{figure}[ht]
    \centering
    \includegraphics[width=1.02\linewidth]{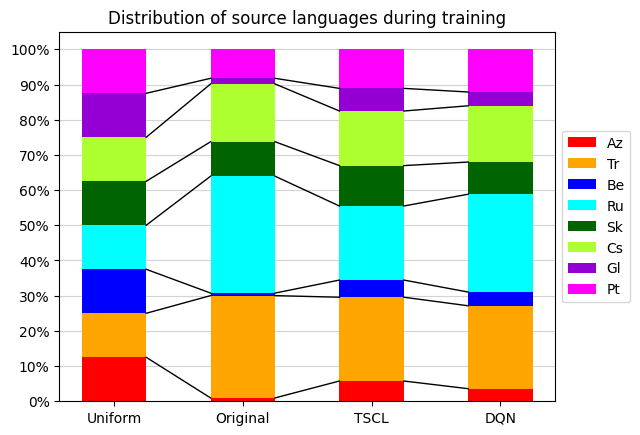}
    \caption{Proportions of data seen during training, as optimized by the TSCL and DQN algorithms, in comparison to uniform (1/8) or original proportions.}
    \label{fig:data-proportions}
\end{figure}

\subsection{Optimized Training Schedule}
\label{sec:training-schedule} 

We claim that the improved average scores with respect to the baselines are due to an optimized training schedule, which can be observed by considering the total amount of data from each language seen during training, shown in Figure~\ref{fig:data-proportions}.  The `uniform' and `proportional' baselines are shown in the first two columns.  In the first case, the NMT model likely overfits to the LRL data, which is seen too often (12.5\% of the times per language) with respect to its diversity (ca.\ 1\% for three LRLs).  In the second case, the number of times each LRL batch is seen during training is insufficient.

Our two algorithms strike a balance between these two extremes, as they are able to automatically determine more suitable proportions of batches of LRLs vs.\ HRLs for training.  We see in Figure~\ref{fig:data-proportions}, third column, how the proportions of three LRLs are increased by TSCL (\textsc{Az} in dark orange, \textsc{Be} in dark blue, and \textsc{Gl} in dark purple).  Two other languages with similar original proportions (\textsc{Pt} in light purple and \textsc{Sk} in dark green) see their proportions increased too, though less than the previous ones.  Conversely, the proportions of HRLs decrease, especially for \textsc{Ru} and \textsc{Tr}.

In comparison to TSCL, the DQN algorithm appears to reach a slightly smaller proportion of LRLs, as seen in the fourth column of Figure~\ref{fig:data-proportions}, where proportions of the darker colors are shrunk with respect to the third column.  These proportions are found quite quickly during training, as can be seen from Figure~\ref{fig:dqn-learned-policy} in Appendix~\ref{app:learned-policies}, where we aggregate the proportions of actions every 1000 steps.


\begin{table*}[ht!]
    \centering
    \begin{tabular}{l|cc|cc|cc|cc}
        \hline
            \textbf{Hyperparameter} & \multicolumn{2}{c|}{\textsc{Az} $\rightarrow$ \textsc{En}} & \multicolumn{2}{c|}{\textsc{Be} $\rightarrow$ \textsc{En}} & \multicolumn{2}{c|}{\textsc{Sk} $\rightarrow$ \textsc{En}} & \multicolumn{2}{c}{\textsc{Gl} $\rightarrow$ \textsc{En}} \\
            \textbf{values} & \textsc{Bleu} & \textsc{Comet} & \textsc{Bleu} & \textsc{Comet} &  \textsc{Bleu} & \textsc{Comet}  & \textsc{Bleu} & \textsc{Comet}  \\    
        \hline
            {Default} & \textbf{14.89} & \textbf{65.04} & \textbf{20.10} & \textbf{61.96} & \textbf{34.35} & \textbf{76.23} & \textbf{32.64} & \textbf{73.59} \\ 
            {Warm-up LRL+HRL} & 14.33 & 64.01 & 19.93 & 61.86 & 33.21 & 75.80 & 31.83 & 72.59 \\
            {No warm-up} & 14.50 & 64.36 & 19.41 & 61.02 & 33.34 & 75.46 & 31.75 & 72.17 \\
            {$\alpha = 0.3$} & 14.28 & 64.72 & 19.71 & 61.74 & 33.51 & 75.68 & 31.79 & 72.56 \\
            {$\epsilon = 0.3$} & 14.10 & 64.37 & 19.95 & 61.80 & 33.36 & 75.75 & 31.16 & 72.38 \\
            {$n = 10$} & 14.74 & 64.92 & 19.83 & 61.30 & 33.66 & 75.93 & 31.71 & 72.82 \\
	\hline
    \end{tabular}
    \caption{MT performance using the TSCL algorithm when hyperparameters vary.}
    \label{tab:tscl-results}
\end{table*}

\subsection{Role of Hyperparameters} 
\label{sec:parameter-setting}

In this section, we study the influence of hyperparameters on the scores of NMT systems trained with the TSCL and DQN algorithms.  We present the scores obtained with significant variations of one parameter at a time in Table~\ref{tab:tscl-results} for TSCL and in Table~\ref{tab:dqn-results} for DQN.  Globally, the scores of the algorithms do not vary much, which shows that they are robust with respect to the variations of the hyperparameters, but also confirms that the algorithms behave consistently from run to run.  For both algorithms, BLEU and COMET scores lead to similar rankings.

For TSCL, we observe first that adding LRLs during warm-up (with uniform frequencies), or skipping warm-up entirely (thus starting with the highest learning rate), are not good options (second and third lines of Table~\ref{tab:tscl-results}).  Instead, cross-lingual transfer from HRLs to LRLs becomes fully beneficial only with a 16k step warm-up on HRLs.  Moreover, convergence is twice slower without warm-up. 
The smoothing coefficient $\alpha$ can vary around the default value of 0.3 with a small decrease in performance ($\alpha = 0.1$ is shown in the 4th line) and so can $\epsilon$ for the $\epsilon$-greedy policy ($\epsilon = 0.3$ instead of 0.1 is shown in the 5th line).  Finally, whether an action is selected every 10 steps or at every step results in comparable scores.

For DQN, we examine first if the Q network is over-parameterized, by reducing the size of the two hidden layers from 512 to 128 (2nd line of Table~\ref{tab:dqn-results}).  This brings only a moderate decrease in average scores, but slightly better COMET scores for \textsc{Az}, \textsc{Sk} and \textsc{Gl}.   

If we vary $\tau$, the smoothing rate of the updates of the Q network (see line 20 of Algorithm~\ref{alg:DQN}) within a large range between 0 and 1, the scores remain stable or even increase for some LRLs (3rd and 4th lines of Table~\ref{tab:dqn-results}, values of 0.5 and 0.995 with respect to default of 0.005).  

Similarly, if we vary $\gamma$, the discount factor for the importance of future rewards, within a large interval between 0 and 1, the scores also remain stable (5th and 6th lines of Table~\ref{tab:dqn-results}, values of 0.5 and 0.01 with respect to default of 0.99).  In fact, the value with the lowest scores is $\gamma = 0.01$, i.e.\ a system that gives only a marginal importance to long-term rewards.  Conversely, this is also the system with the fastest convergence, although no particular variant seems to be particularly slow to converge (see Table~\ref{tab:dqn-checkpoints} in Appendix~\ref{app:convergence-speed}), and none achieves highest scores on all LRLs.

We can thus conclude that the default values of hyperparameters of TSCL and DQN used in Section~\ref{sec:translation-performance} above, inspired respectively by \citet{Matiisen-2017} and by \citet{kumar-etal-2019-reinforcement}, perform well and that both algorithms are stable when these hyperparameters vary.

\begin{table*}[ht!]
\centering
\begin{tabular}{l|cc|cc|cc|cc}
\hline
\textbf{Hyperparameter} & \multicolumn{2}{c|}{\textsc{Az} $\rightarrow$ \textsc{En}} & \multicolumn{2}{c|}{\textsc{Be} $\rightarrow$ \textsc{En}} & \multicolumn{2}{c|}{\textsc{Sk} $\rightarrow$ \textsc{En}} & \multicolumn{2}{c}{\textsc{Gl} $\rightarrow$ \textsc{En}} \\
\textbf{values} & \textsc{Bleu} & \textsc{Comet} & \textsc{Bleu} & \textsc{Comet} &  \textsc{Bleu} & \textsc{Comet}  & \textsc{Bleu} & \textsc{Comet}  \\    
\hline
{Default} & 15.62 & 65.86 & \textbf{21.11} & \textbf{62.82} & 34.54 & 76.15 & 33.02 & 73.73 \\
{Hidden size $=128$} & 15.86 & \textbf{66.15} & 20.38 & 62.64 & 34.40 & 76.21 & 32.59 & 73.79 \\
{$\tau=0.5$} & 15.55 & 66.13 & 20.17 & 62.47 & 34.52 & 76.25 & \textbf{33.13} & \textbf{73.81} \\
{$\tau=0.995$} & \textbf{16.06} & 66.06 & 20.19 & 62.18 & \textbf{34.55} & \textbf{76.31} & 32.45 & 73.46 \\
{$\gamma=0.5$} & 15.78 & 65.85 & 20.51 & 62.64 & 34.49 & 76.16 & 32.94 & 73.63 \\
{$\gamma=0.01$} & 15.34 & 65.38 & 20.59 & 62.36 & 33.87 & 75.70 & 32.43 & 73.39 \\
\hline
\end{tabular}
\caption{MT performance using the DQN algorithm when hyperparameters vary.}
\label{tab:dqn-results}
\end{table*}


\subsection{Analysis of the Q Network}
\label{sec:q-network}

In this section, we propose a method to analyze the Q network of the DQN algorithm, which predicts on what language to train next, given a vector of 200 scores of an NMT model.  Specifically, these scores are the cross-entropy loss values on 200 monolingual batches of 10 sentences each from the prototype set.  At a given moment during training, the Q network can be probed with a specific vector as input, for instance a vector that represents a specific state of the NMT system.  We propose to probe the Q network with a state in which one language is poorly learned.  This is mimicked by assigning high loss values to the coefficients of the vector that represent scores on batches of this language.  To avoid an entirely synthetic vector, we pick an actual vector occurring during training and multiply by 5 the loss values of all 25 batches from the targeted  language. 

\begin{figure}[ht]
    \centering
    \includegraphics[width=0.98\linewidth]{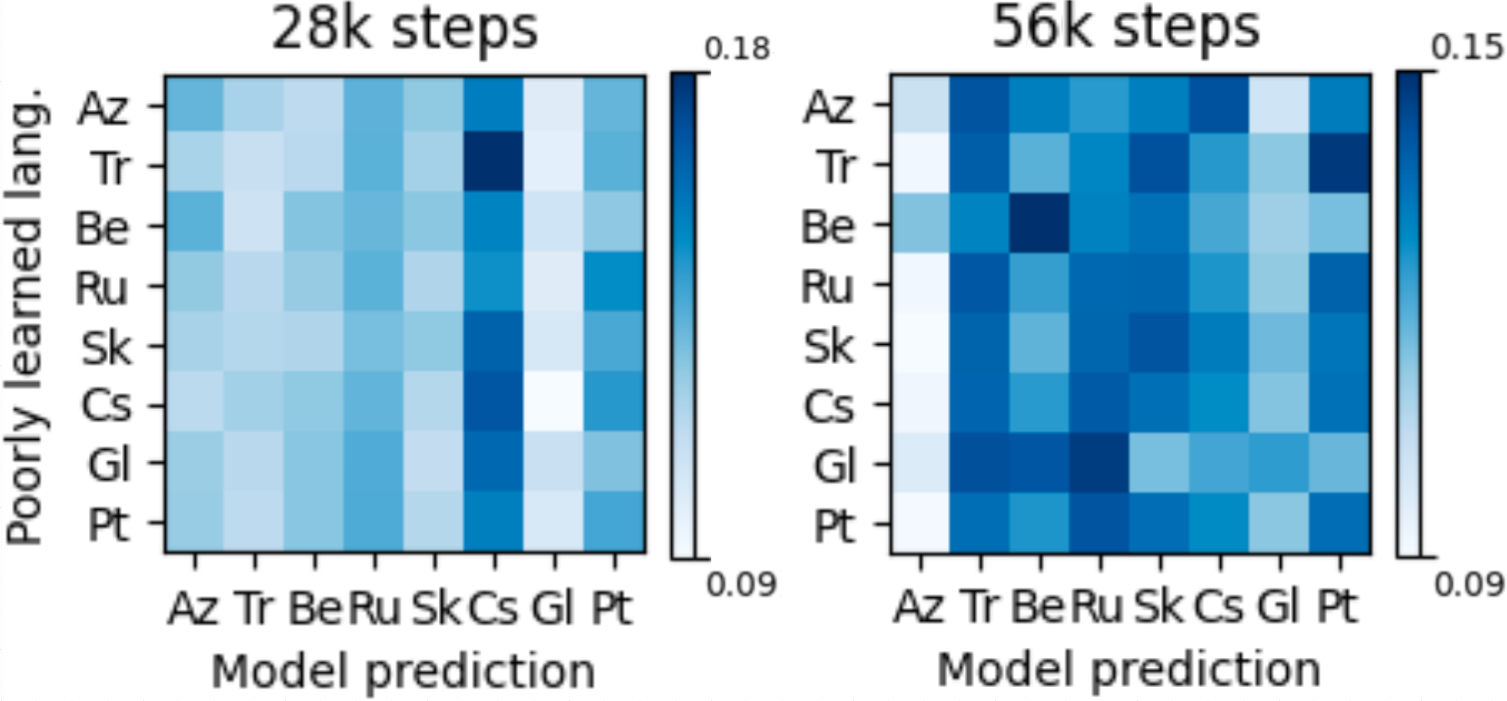}
    \caption{Behavior of the Q network at two stages during training.}
    \label{fig:model-predictions-28k-56k}
\end{figure}

We probe the Q network with each of the 8 source languages in turn, pretending that this language is not well learned and observing the action selected by the network, i.e.\ the language that it requires the NMT model to see next.  Rather than observing the single selected language, we considered the softmaxed output activations for all 8 languages. The result is thus an 8-by-8 matrix, represented in Figure~\ref{fig:model-predictions-28k-56k} at 28k and respectively 56k training steps.  The X-axis represents the softmaxed output activations (predicted Q-values for each language), while the lines of the Y-axis correspond to each probed language (the one for which loss values were amplified).

The state at 28k steps is typical of incomplete training.  The Q network appears to favor one HRL language (\textsc{Cz}) regardless of the language that is the least well learned according to its synthetic vector.  The Q network selects one language for a large number of steps and gradually switches to another.  The state at 56k steps (when the NMT trained with DQN reaches its best score) demonstrates a more balanced behavior: if one language is insufficiently learned, especially a LRL, then the network predicts that more training should be done on that language.  Indeed, for several lines (though not all), the cell on the diagonal is one of the darkest of the line (e.g.\ for \textsc{Tr}, \textsc{Be}, \textsc{Sk} or \textsc{Pt}).  These observations suggest that the DQN model's decisions are complex and evolve over time, rather than always favoring the language that is currently the least well learned.


\section{Conclusion and Perspectives}

In this paper, we presented two algorithms for optimizing the training schedule of multilingual NMT models when a mixture of HRLs and LRLs must be learned on the source side.  The TSCL algorithm models the expected return of each action by smoothing past observations, while DQN trains a neural network to perform this estimation and to select the optimal action.  

Both algorithms strike a balance between a uniform distribution of training batches across languages and a distribution purely based on the respective frequencies of these languages in the actual data.  The algorithms increase the proportions of LRLs and reduce those of HRLs, while still enabling cross-lingual transfer from HRLs to related LRLs.  The better balance of HRLs and LRLs avoids too great a focus on the more abundant HRL data (which would sacrifice LRLs) or too great a focus on LRLs (which would lead to overfitting).  Without such algorithms, it would be difficult to find extrinsic criteria to optimize the presentation frequencies of batches. Moreover, the optimized training schedules lead to improved macro-average BLEU and COMET scores.

We leave for future work the study of other ways to construct batches.  One option is to use multilingual batches -- though, as shown above, shuffled batches underperform with respect to an optimized balance of LRLs and HRLs.  Another option is to define actions as specific batches or groups of batches, which would enable the model to prioritize certain batches over others, but would also increase the number of possible actions and hence the learning complexity of the RL agent.

The relevance of our proposal should be tested with additional datasets combining HRLs and related LRLs, and with other neural architectures for which cross-lingual transfer may be important to ensure acceptable performance on LRLs, particularly LLMs fine-tuned on translation tasks \citep{alves2024toweropenmultilinguallarge}.  In such cases, an optimized training schedule across available resources may also be beneficial.

\section*{Acknowledgments}
We are grateful for its support to the Swiss National Science Foundation, through the DOMAT grant n.\ 175693, On-demand Knowledge for Document-level Machine Translation and the EXOMAT grant n.\ 228494, External Knowledge for Low-resource Machine Translation.  We thank Giorgos Vernikos for feedback on earlier versions of the paper.

\bibliography{anthology,custom}

\clearpage
\input{appendix}

\end{document}

%% file: appendix.tex
\appendix

\section{Appendix}
\label{sec:appendix}

\subsection{Epsilon Scheduling for DQN}
\label{app:epsilon-scheduling}

\begin{figure}[ht]
    \centering
    \includegraphics[width=0.8\linewidth]{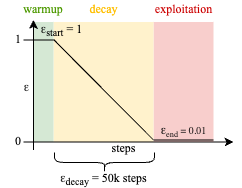}
    \caption{Evolution of $\epsilon$ during training with DQN.}
    \label{fig:epsilon-decay}
\end{figure}

The RL agent in the DQN algorithm follows, as explained in Section~\ref{sec:dqn-algorithm}, an $\epsilon$-greedy policy: with a probability of $1 - \epsilon$, actions (i.e.\ the source language of a batch) are selected using the Q network, but with a probability of $\epsilon$, a random action is selected.  The exact schedule of $\epsilon$ is shown in Figure~\ref{fig:epsilon-decay}.  

During the warm-up period of 16k a value of 1 means that the Q network is not used and source HRLs (in this case) are drawn randomly.  Then, during a decay of 50k steps, the importance of the Q network in deciding the actions grows progressively, while random choices decrease to a minimal probability of 0.01 after 66k steps.  This approach, inspired by \citet{kumar-etal-2019-reinforcement}, achieves a balance between exploiting the Q network and exploring new actions.

\subsection{Convergence Speed}
\label{app:convergence-speed}

In the experiments presented above, the scores were computed using a rolling ensemble of 4 checkpoints, and the best score was selected as the highest macro-average of BLEU achieved on the development data of the LRLs.  We mentioned at the end of Section~\ref{sec:translation-performance} that the DQN was the method that reached optimal scores after the smallest number of steps, followed by TSCL and then by the `proportional' scheduling.  The exact numbers of steps are given in Table~\ref{tab:checkpoints}, showing that DQN accelerates convergence with respect to the other schedules.  Moreover, this behavior is stable when varying some of the hyperparameters of the algorithm, as shown in Table~\ref{tab:dqn-checkpoints}. 

\begin{table}[ht] 
    \centering
    \begin{tabular}{ll|c}
        \toprule
            \textbf{Training} & \textbf{Warm up} & \textbf{Best checkpoint} \\
        \midrule
            Uniform & no & 136k\\
            Proportional & no & 60k\\
            Uniform & 16k & 124k\\
            Proportional & 16k & 60k\\
            Shuffle batch & 16k & 128k\\
            TSCL & 16k & 56k\\
            DQN & 16k & 52k\\
        \bottomrule
    \end{tabular}
    \caption{Comparison of the number of steps required by the NMT model to achieve the best scores on the LRLs.}
    \label{tab:checkpoints}
\end{table}

\begin{table}[ht]
    \centering
    \begin{tabular}{l|c}
        \toprule
            \textbf{Parameter values} & \textbf{Best checkpoint} \\
        \midrule
            {Default} & 52k\\
            {$\tau=0.5$} & 60k\\
            {$\tau=0.995$} & 76k\\
            {$\gamma=0.5$} & 48k\\
            {$\gamma=0.01$} & 36k\\
            {Hidden layer: $128$} & 76k\\
        \bottomrule
    \end{tabular}
    \caption{Comparison of the number of steps required for the NMT model using the DQN algorithm to achieve the best scores on the LRLs.  The parameter values are the default ones, except the changes shown for each line.}
    \label{tab:dqn-checkpoints}
\end{table}

\subsection{Learned Policies}
\label{app:learned-policies}

\begin{figure*}[ht]
    \centering
    \includegraphics[width=\textwidth]{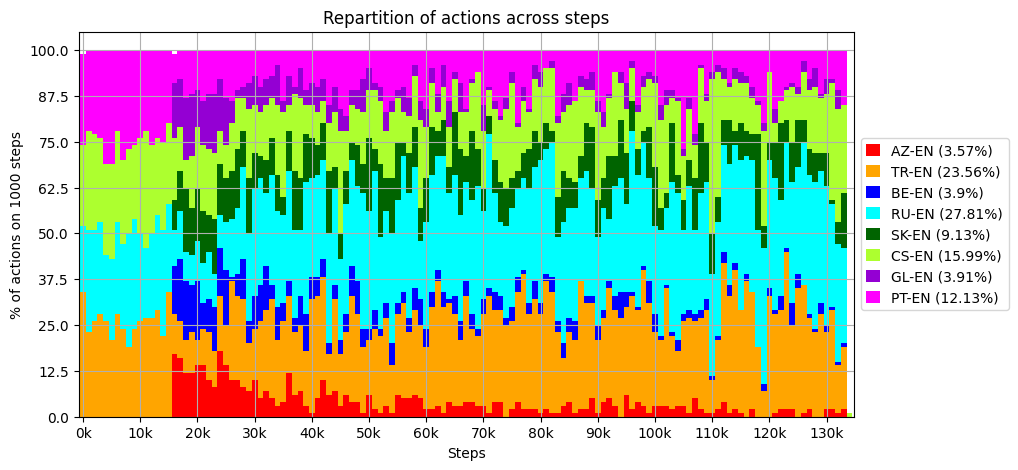}
    \caption{Evolution of the proportions of actions (i.e.\ language of batches) chosen during the training of the DQN system. Aggregation windows of 1000 steps are computed. The first 16k steps are the warmup on the HRLs only.\vspace{1em}}
    \label{fig:dqn-learned-policy}
\end{figure*}

We presented in Section~\ref{sec:training-schedule} the total numbers of actions of each type (i.e.\ source language of the batch) selected during training for the TSCL and DQN algorithms, in comparison to the `uniform' and `proportional' training schedules.  Here, we show in Figure~\ref{fig:dqn-learned-policy} the evolution of the proportion of actions during training with the DQN algorithm, aggregated every 1000 steps. 

In this representation, we first observe that the initial 16k steps are performed only on the HRLs, as configured. When the DQN algorithm starts playing a role, a random selection of languages is observed.  As the algorithm learns, the proportions of LRLs decrease, while the proportions of HRLs increase, and tend to stabilize towards steady-state values. The proportions averaged over the entire training period are provided in the legend of the chart. These are the proportions compared between the systems in Figure~\ref{fig:data-proportions}.


\subsection{The TSCL Algorithm}
\label{sec:appendix-tscl}

The full specification of the TSCL algorithm in pseudo-code is provided hereafter as Algorithm~\ref{alg:TSCL_online}.

\begin{algorithm*}[ht]
    \DontPrintSemicolon
    \SetKwInput{Input}{Input}
    \SetKwInput{Require}{Require}
    \SetKwInOut{Output}{Output}
    \Require{actions $\mathcal{A}\leftarrow\{A_1, .., A_k\}$, number of training steps $ts$, number of consecutive actions $n$, number of warm-up steps $w$, $\epsilon$-greedy policy exploration parameter $\epsilon$, smoothing coefficient $\alpha$}
    Initialize NMT model\;
    Initialize action index $i \leftarrow1$\;
    Initialize unvisited actions indexes $U\leftarrow \{2, .., k\}$\;
    Initialize estimated return $Q(A_k)\leftarrow0$ for all $k$ actions\;
    Initialize rewards history $H(A_k)\leftarrow0$ for all $k$ actions\;
    \For{$t \gets 1, \ldots, ts$}
    {
        Sample mini-batch $B_t$ from action $A_i$\;
        Train NMT model using mini-batch $B_t$\;
        \If{$t \bmod n = 0$}{
            Observe reward $R_t \leftarrow X_t - H(A_i)$\;
            Update reward history $H(A_i) \leftarrow X_t$\;
            Exponentially smooth estimated return $Q(A_i) \leftarrow \alpha R_t + (1-\alpha)Q(A_i)$\;
            
            \If{$|U| \neq 0$}{
                Choose action index $i \leftarrow U[0]$\;
                Update $U \leftarrow U - \{i\}$\;
            }
            \Else{
                Choose random number $r$ between 0 and 1\;
                \If{$t < w$ \textbf{or} $r < \epsilon$}{
                    Choose action index $i$ randomly between 1 and $k$\;
                }
                \Else{
                    Set $i$ as the index of the max arg.\ of absolute values in $Q$\;
                }
            }
        }
    }
    \caption{TSCL algorithm for NMT.}
    \label{alg:TSCL_online}
\end{algorithm*}

\subsection{The DQN Algorithm}
\label{sec:appendix-dqn}

The full specification of the DQN algorithm in pseudo-code is provided hereafter as Algorithm~\ref{alg:DQN}.

\begin{algorithm*}[ht]
    \DontPrintSemicolon
    \SetKwInput{Input}{Input}
    \SetKwInput{Require}{Require}
    \SetKwInOut{Output}{Output}
    \Require{
    actions $\mathcal{A}\leftarrow\{A_1, .., A_k\}$, number of training steps $ts$, number of consecutive actions $n$, number of warm-up steps $w$, $\epsilon$-greedy policy exploration parameter $\epsilon$, soft update coefficient $\tau$, replay memory capacity $c$, minimum replay memory capacity $c_{min}$
    }
    Initialize NMT model learning algorithm\;
    Initialize RL agent's online model\;
    Initialize replay memory deque $\mathcal{D}$ with capacity $c$\;
    Initialize RL agent's target network with same weights as RL agent's online model\;
    Initialize action index $i \leftarrow1$\;
    \For{$t \gets 1, \ldots, T$}
    {
        Sample mini-batch $B_t$ from action $A_i$\;
        Train NMT model using mini-batch $B_t$\;
        \If{$t \bmod n = 0$}{
            \If{$t < w$}{
                Choose action index $i$ randomly between 1 and $k$\;
            }
            \Else{
                Observe current state $S_t$\;
                Observe reward $R_t \leftarrow X_t - X_{t-1}$\;
                Store transition $(S_{t-1}, i, R_t, S_t)$ in replay memory $\mathcal{D}$\;
                \If{$|D|>=c_{min}$}{
                    Sample mini-batch of transitions $T$ from replay memory $\mathcal{D}$\;
                    Train RL agent's online model using mini-batch $T$\;
                    Soft update RL agent's target model weights with RL agent's online model weights $\theta^- \leftarrow \tau \theta + (1 - \tau) \theta^-$\;
                }
                Choose random number $r$ between 0 and 1\;
                \If{$r < \epsilon$}{
                    Choose action index $i$ randomly between 1 and $k$\;
                }
                \Else{
                    Predict Q values at state $S_t$ with RL agent target network\;
                    Set $i$ as the index of the arg. max in $Q$\;
                }
                Decrease $\epsilon$ according to decay schedule\;
            }
        }
    }
    \caption{DQN Algorithm for NMT}
    \label{alg:DQN}
\end{algorithm*}